\documentclass[11pt]{article}
\usepackage{emnlp2015}
\usepackage{times}
\usepackage{url}
\usepackage{latexsym}
\usepackage{epsfig}
\usepackage{array}
\usepackage{color}
\usepackage{multirow}
\usepackage{url}
\usepackage{subfig}
\usepackage{algorithmic}
\usepackage{algorithm}
\usepackage{multirow}
\usepackage{rotating}
\usepackage{booktabs}
\usepackage{makecell}
\usepackage{comment}
\usepackage{amsmath,amsfonts,amssymb}
\usepackage{tikz}
\usepackage{verbatim}
\usepackage{caption}
\usepackage{booktabs}
\usepackage{ctable}
\usepackage{makecell}
\usepackage{array} 
\usepackage{varwidth} 

\newcommand*{\Scale}[2][4]{\scalebox{#1}{$#2$}}%

\newcolumntype{M}{>{\begin{varwidth}{6cm}}l<{\end{varwidth}}} 
\newcolumntype{C}[1]{>{\centering\let\newline\\\arraybackslash\hspace{0pt}}m{#1}}
\newcolumntype{L}[1]{>{\raggedright\let\newline\\\arraybackslash\hspace{0pt}}m{#1}}

\title{Semantically Conditioned LSTM-based Natural Language Generation for Spoken Dialogue Systems}

\author{Tsung-Hsien Wen, Milica Ga{\v{s}}i\'c, Nikola Mrk{\v{s}}i\'c, \\ {\bf Pei-Hao Su, David Vandyke \and  Steve Young} \\
  Cambridge University Engineering Department,  \\
  Trumpington Street, 
  Cambridge, CB2 1PZ, UK\\
  {\tt \{thw28,mg436,nm480,phs26,djv27,sjy\}@cam.ac.uk}\
 }

\date{}
\begin{document}
\maketitle

\begin{abstract}

Natural language generation (NLG) is a critical component of spoken dialogue and it has a significant impact both on usability and perceived quality.  Most NLG systems in common use employ rules and heuristics and tend to generate rigid and stylised responses without the natural variation of human language.  They are also not easily scaled to systems covering multiple domains and languages.
This paper presents a statistical language generator based on a semantically controlled Long Short-term Memory (LSTM) structure.
The LSTM generator can learn from unaligned data by jointly optimising sentence planning and surface realisation  using a simple cross entropy training criterion, and language variation can be easily achieved by sampling from output candidates.   
With fewer heuristics, an objective evaluation in two differing test domains showed the proposed method improved performance compared to previous methods. 
Human judges scored the LSTM system higher on informativeness and naturalness and overall preferred it to the other systems.

\end{abstract}

\section{Introduction}
\label{sec:intro}

The natural language generation (NLG) component provides much of the persona of a spoken dialogue system (SDS), and it has a significant impact on a user's impression of the system.
As noted in \newcite{Stent05evaluatingevaluation}, a good generator usually depends on several factors: adequacy, fluency, readability, and variation.
Previous approaches attacked the NLG problem in different ways. 
The most common and widely adopted today is the {\it rule-based} (or {\it template-based}) approach \cite{cheyer2007method,mirkovic2011dialogue}.
Despite its robustness and adequacy, the frequent repetition of identical, rather stilted, output forms make talking to a {\it rule-based} generator rather tedious.  Furthermore, the approach does not easily scale to large open domain systems\cite{6407655,gavsic2014incremental,Henderson2014d}.  Hence approaches to NLG are required that can be readily scaled whilst meeting the above requirements.

The {\it trainable generator} approach exemplified by the HALOGEN \cite{Langkilde1998} and  SPaRKy system \cite{Stent04trainablesentence} provides a possible way forward.
These systems include specific trainable modules within the generation framework to allow the model to adapt to different domains \cite{Walker07individualand}, or reproduce certain style \cite{Mairesse2011CUP}.
However, these approaches still require a handcrafted generator to define the decision space within which statistics can be used for optimisation.  
The resulting utterances are therefore constrained by the predefined syntax and any domain-specific colloquial responses must be added manually.

More recently, {\it corpus-based} methods \cite{Oh2000,Mairesse2014,wenrgm15} have received attention as access to  data becomes increasingly available.
By defining a flexible learning structure, {\it corpus-based} methods aim to learn generation directly from data by adopting an over-generation and reranking paradigm \cite{Oh2000}, in which final responses are obtained by reranking a set of candidates generated from a stochastic generator.
Learning from data directly enables the system to mimic human responses more naturally, removes the dependency on predefined rules, and makes the system easier to build and extend to other domains.
As detailed in Sections~\ref{sec:related} and~\ref{sec:neural_gen}, however, these existing approaches have weaknesses in the areas of training data efficiency, accuracy and naturalness.


This paper presents a statistical NLG based on a semantically controlled Long Short-term Memory (LSTM) recurrent network.
It can learn from unaligned data by jointly optimising its sentence planning and surface realisation components using a simple cross entropy training criterion without any heuristics,  and good quality language variation is obtained simply  by randomly sampling the network outputs.
We start in Section \ref{sec:neural_gen} by defining the framework of the proposed neural language generator.
We introduce the semantically controlled LSTM (SC-LSTM) cell in Section \ref{subsec:sclstm}, then we discuss how to extend it to a deep structure in Section \ref{subsec:deep}.
As suggested in \newcite{wenrgm15}, a backward reranker is introduced in Section \ref{sec:brnn} to improve fluency.
Training and decoding details are described in Section \ref{sec:train} and \ref{sec:decode}.

Section \ref{sec:exp} presents an evaluation of the proposed approach in the context of an application providing information about venues in the San Francisco area. 
In Section \ref{sec:objective}, we first show that our generator outperforms several baselines using objective metrics.
We experimented on two different ontologies to show not only that good performance can be achieved across domains, but how easy and quick the development lifecycle is.
In order to assess the subjective performance of our system, a quality test and a pairwise preference test are presented in Section \ref{sec:human_eval}. 
The results show that our approach can produce high quality utterances that are considered to be more natural and are preferred to previous approaches.
We conclude with a brief summary and future work in Section \ref{sec:conclusion}.

\section{Related Work}
\label{sec:related} 

Conventional approaches to NLG typically divide the task into sentence planning and surface realisation.
Sentence planning maps input semantic symbols into an intermediary form representing the utterance, e.g. a tree-like or template structure, then surface realisation converts the intermediate structure into the final text \cite{walker2002training,Stent04trainablesentence}.
Although statistical sentence planning has been explored previously, for example, 
generating the most likely context-free derivations given a corpus \cite{Belz2008} or maximising the expected reward using reinforcement learning \cite{Rieser2010}, these methods still rely on a pre-existing, handcrafted generator.
To minimise handcrafting, \newcite{Stent2009} proposed learning sentence planning rules directly from a corpus of utterances labelled with Rhetorical Structure Theory (RST) discourse relations \cite{mann88b}.
However, the required corpus labelling is expensive and additional handcrafting is still needed to map the sentence plan to a valid syntactic form.

As noted above, {\it corpus-based} NLG aims at learning generation decisions from data with minimal dependence on rules and heuristics.
A pioneer in this direction is the class-based n-gram language model (LM) approach proposed by \newcite{Oh2000}.
\newcite{Ratnaparkhi2002435} later addressed some of the limitations of class-based LMs in the over-generation phase by using a modified generator based on a syntactic dependency tree.
\newcite{Mairesse2014} proposed a phrase-based NLG system based on factored LMs that can learn from a semantically aligned corpus. 
Although active learning \cite{Mairesse2010} was also proposed to allow learning  online directly from users, the requirement for human annotated alignments limits the scalability of the system.
Another similar approach casts NLG as a template extraction and matching problem, e.g., \newcite{Angeli2010} train a set of log-linear models to make a series of generation decisions to choose the most suitable template for realisation. 
\newcite{kondadadi13} later show that the outputs can be further improved by an SVM reranker making them comparable to human-authored texts. 
However, template matching approaches do not generalise well to unseen combinations of semantic elements.

The use of neural network-based (NN) approaches to NLG is relatively unexplored.
The stock reporter system ANA by \newcite{Kukich1987} is perhaps the first NN-based generator, although generation was only done at the phrase level.
Recent advances in recurrent neural network-based language models (RNNLM) \cite{39298195,5947611} have demonstrated the value of distributed representations and the ability to model arbitrarily long dependencies.
\newcite{ICML2011Sutskever_524} describes a simple variant of the RNN that can generate meaningful sentences by learning from a character-level corpus.
More recently, \newcite{KarpathyF14} have demonstrated that an RNNLM is capable of generating image descriptions by conditioning the network model on a pre-trained convolutional image feature representation.
\newcite{zhangEMNLP2014} also describes interesting work using RNNs to generate Chinese poetry.
A forerunner of the system presented here is described in \newcite{wenrgm15}, in which a forward RNN generator, a CNN reranker, and a backward RNN reranker are trained jointly to generate utterances.
Although the system was easy to train and extend to other domains, a heuristic gate control was needed to ensure that all of the attribute-value information in the system's response was accurately captured by the generated utterance.
Furthermore, the handling of unusual slot-value pairs by the CNN reranker was rather arbitrary.
In contrast, the LSTM-based system described in this paper can deal with these problems automatically by learning the control of gates and surface realisation jointly.

Training an RNN with long range dependencies is difficult because of the vanishing gradient problem \cite{279181}.
\newcite{Hochreiter1997} mitigated this problem by replacing the sigmoid activation in the RNN recurrent connection with a self-recurrent memory block and a set of multiplication gates to mimic the read, write, and reset operations in digital computers.
The resulting architecture is dubbed the Long Short-term Memory (LSTM) network.
It has been shown to be effective in a variety of tasks, such as speech recognition \cite{DLSTM13}, handwriting recognition \cite{4531750}, spoken language understanding \cite{export228844}, and machine translation \cite{SutskeverVL14}.
Recent work by \newcite{GravesWD14} has demonstrated that an NN structure augmented with a carefully designed memory block and differentiable read/write operations can learn to mimic computer programs. 
Moreover, the ability to train deep networks provides a more sophisticated way of exploiting relations between labels and features, therefore making the prediction more accurate \cite{6296526}.
By extending an LSTM network to be both deep in space and time, \newcite{Graves13} shows the resulting network can used to synthesise handwriting indistinguishable from that  of a human. 

\section{The Neural Language Generator}
\label{sec:neural_gen}

The generation model proposed in this paper is based on a recurrent NN architecture \cite{39298195} in which a 1-hot encoding $\mathrm{\mathbf{w}}_{t}$ of a token\footnote{
We use {\it token} instead of {\it word} because our model operates on text for which slot values are replaced by its corresponding slot tokens. We call this procedure delexicalisation.} 
$w_t$ is input at each time step $t$ conditioned on a recurrent hidden layer $\mathrm{\mathbf{h}}_t$
and outputs the probability distribution of the next token ${w}_{t+1}$.
Therefore, by sampling input tokens one by one from the output distribution of the RNN until a stop sign is generated \cite{KarpathyF14} or some constraint is satisfied \cite{zhangEMNLP2014}, the network can produce a sequence of tokens which can be lexicalised \footnote{The process of replacing slot token by its value.} to form
the required utterance.

\subsection{Semantic Controlled LSTM cell}
\label{subsec:sclstm}
\begin{figure}[h]
\centerline{\includegraphics[width=70mm]{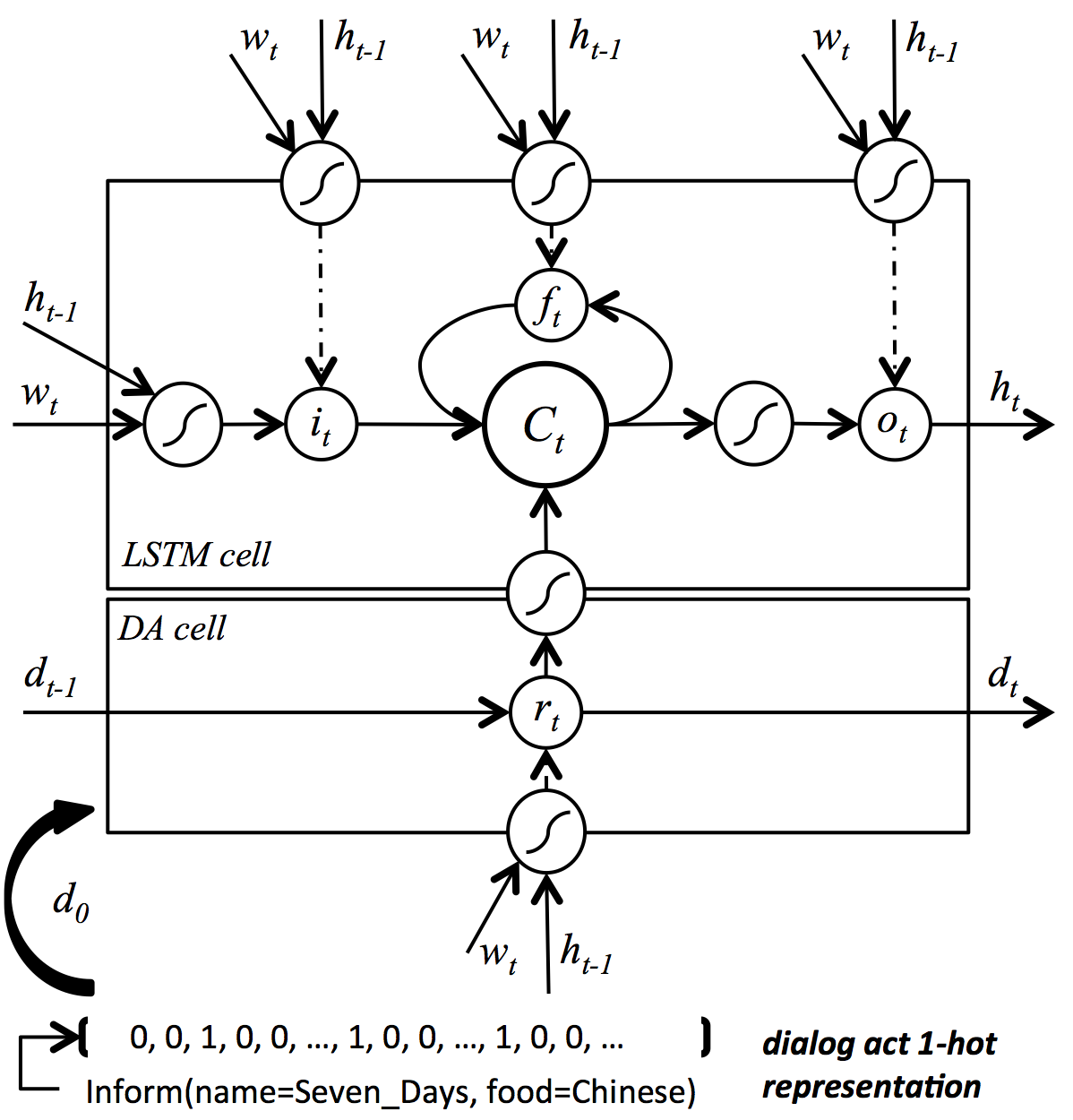}}
\caption{Semantic Controlled LSTM cell proposed in this paper. The upper part is a traditional LSTM cell in charge of surface realisation, while the lower part is a sentence planning cell based on a sigmoid control gate and a dialogue act (DA). }
\label{fig:lstm}
\end{figure}
Long Short-term Memory \cite{Hochreiter1997} is a recurrent NN architecture which uses a vector of memory cells $\mathrm{\mathbf{c}}_t \in \mathbb{R}^{n}$ and a set of elementwise multiplication gates to control how information is stored, forgotten, and exploited inside the network.
Of the various different connectivity designs for an LSTM cell \cite{Graves13,ZarembaSV14}, the architecture used in this paper is illustrated in Figure~\ref{subsec:sclstm} and defined by the following equations,,
\begin{alignat}{6}
\mathrm{\mathbf{i}}_t &= \sigma( \mathrm{\mathbf{W}}_{wi}\mathrm{\mathbf{w}}_t +\mathrm{\mathbf{W}}_{hi}\mathrm{\mathbf{h}}_{t-1} )\label{eq:igate}\\
\mathrm{\mathbf{f}}_t &= \sigma( \mathrm{\mathbf{W}}_{wf}\mathrm{\mathbf{w}}_t +\mathrm{\mathbf{W}}_{hf}\mathrm{\mathbf{h}}_{t-1} )\label{eq:fgate}\\
\mathrm{\mathbf{o}}_t &= \sigma( \mathrm{\mathbf{W}}_{wo}\mathrm{\mathbf{w}}_t +\mathrm{\mathbf{W}}_{ho}\mathrm{\mathbf{h}}_{t-1} )\label{eq:ogate}\\
\hat{\mathrm{\mathbf{c}}}_t &= tanh( \mathrm{\mathbf{W}}_{wc}\mathrm{\mathbf{w}}_t +\mathrm{\mathbf{W}}_{hc}\mathrm{\mathbf{h}}_{t-1} )\label{eq:c_hat}\\
\mathrm{\mathbf{c}}_t &= \mathrm{\mathbf{f}}_{t} \odot \mathrm{\mathbf{c}}_{t-1} + \mathrm{\mathbf{i}}_{t} \odot \hat{\mathrm{\mathbf{c}}}_t\label{eq:c}\\
\mathrm{\mathbf{h}}_t &=  \mathrm{\mathbf{o}}_{t} \odot tanh(\mathrm{\mathbf{c}}_t)\label{eq:h}
\end{alignat}
where $\sigma$ is the sigmoid function, $\mathrm{\mathbf{i}}_t,\mathrm{\mathbf{f}}_t,\mathrm{\mathbf{o}}_t \in [0,1]^n$ are input, forget, and output gates respectively, and $\hat{\mathrm{\mathbf{c}}}_t$ and $\mathrm{\mathbf{c}}_t$ are proposed cell value and true cell value at time $t$. 
Note that each of these vectors has a dimension equal to the hidden layer $\mathrm{\mathbf{h}}$.

In order to ensure that the generated utterance represents the intended meaning, the generator is further conditioned on a control vector $\mathrm{\mathbf{d}}$, a 1-hot representation of the dialogue act (DA) type and its slot-value pairs.
Although a related work \cite{KarpathyF14} has suggested that reapplying this auxiliary information to the RNN at every time step can increase performance by mitigating the {\it vanishing gradient problem} \cite{6424228,279181}, we have found that such a model also omits and duplicates slot information in the surface realisation. 
In \newcite{wenrgm15} simple heuristics are used to turn off slot feature values in the control vector $\mathrm{\mathbf{d}}$ once the corresponding slot token has been generated. However, these heuristics can only handle cases where slot-value pairs
can be identified by exact matching between the delexicalised surface text and the slot value pair encoded in $\mathrm{\mathbf{d}}$.   Cases such as binary slots and slots that take don't care values cannot be explicitly delexicalised in this way and these cases frequently result in generation errors.

\begin{figure*}[t]
\centerline{\includegraphics[width=110mm]{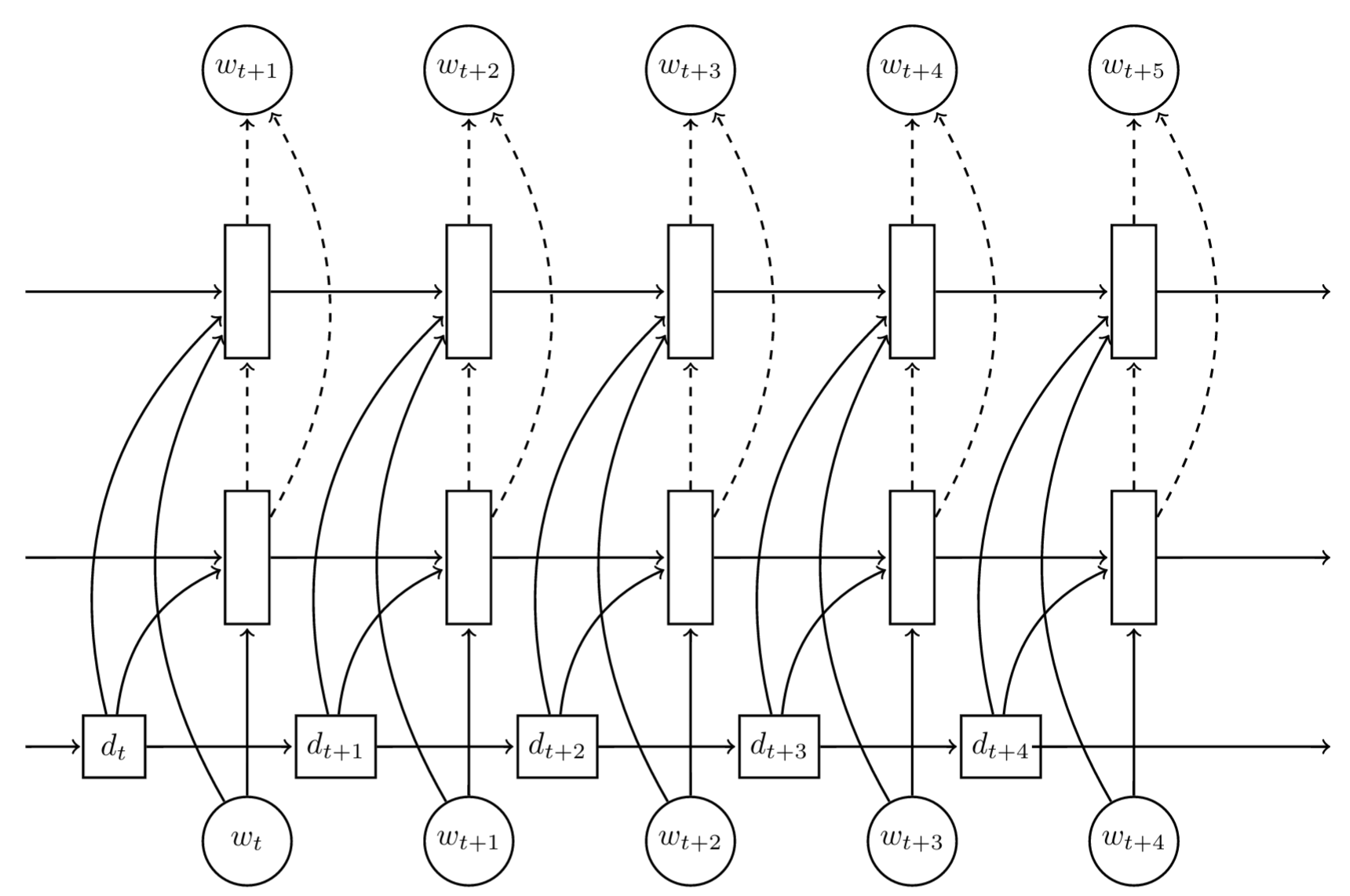}}
\caption{The Deep LSTM generator structure by stacking multiple LSTM layers on top of the DA cell. The skip connection was adopted to mitigate the vanishing gradient, while the dropout was applied on dashed connections to prevent co-adaptation and overfitting. }
\vspace{-2 mm}
\label{fig:deep}
\end{figure*}

To address this problem, an additional control cell is introduced into the LSTM to gate the DA as shown in Figure \ref{fig:lstm}.
This cell plays the role of {\it sentence planning} since it manipulates the DA features during the generation process in order to produce
a surface realisation which accurately encodes the input information.
We call the resulting architecture Semantically Controlled LSTM (SC-LSTM).
Starting from the original DA 1-hot vector $\mathrm{\mathbf{d}}_0$, at each time step the DA cell decides what information should be retained for  future time steps and discards the others,
\begin{alignat}{2}
\mathrm{\mathbf{r}}_t &= \sigma( \mathrm{\mathbf{W}}_{wr}\mathrm{\mathbf{w}}_t +\alpha\mathrm{\mathbf{W}}_{hr}\mathrm{\mathbf{h}}_{t-1} )\label{eq:rgate}\\
\mathrm{\mathbf{d}}_t &= \mathrm{\mathbf{r}}_{t} \odot \mathrm{\mathbf{d}}_{t-1} \label{eq:d}
\end{alignat}
where $\mathrm{\mathbf{r}}_t \in [0,1]^d$ is called the reading gate, and $\alpha$ is a constant. 
Here  $\mathrm{\mathbf{W}}_{wr}$ and $\mathrm{\mathbf{W}}_{hr}$ act like keyword and key phrase detectors that learn to associate certain patterns of generated tokens with certain slots.
Figure ~\ref{fig:example} gives an example of how these detectors work in affecting DA features inside the network.
Equation \ref{eq:c} is then modified so that the cell value $\mathrm{\mathbf{c}}_{t}$ also depends on the DA, 
\begin{equation}
\label{eq:newc}
\mathrm{\mathbf{c}}_t = \mathrm{\mathbf{f}}_{t} \odot \mathrm{\mathbf{c}}_{t-1} + \mathrm{\mathbf{i}}_{t} \odot \hat{\mathrm{\mathbf{c}}}_t + tanh(\mathrm{\mathbf{W}}_{dc}\mathrm{\mathbf{d}}_t)
\end{equation}
After updating Equation \ref{eq:h} by Equation \ref{eq:newc},  the output distribution is formed by applying a softmax function $g$, and the distribution is sampled to obtain the next token, 
\begin{equation}
\Scale[1]{P(w_{t+1}|w_t,w_{t-1},...w_0,\mathrm{\mathbf{d}}_t) = g( \mathrm{\mathbf{W}}_{ho}\mathrm{\mathbf{h}}_t )}
\end{equation}
\begin{equation}
\Scale[1]{w_{t+1} \sim P(w_{t+1}|w_t,w_{t-1},...w_0,\mathrm{\mathbf{d}}_{t})}.
\end{equation}

\subsection{The Deep Structure}
\label{subsec:deep}

Deep Neural Networks (DNN) enable increased discrimination by learning multiple layers of features, and represent  the state-of-the-art for many applications  such as speech recognition \cite{DLSTM13} and natural language processing \cite{Collobert2008}.
The neural language generator proposed in this paper can be easily extended to be deep in both space and time by stacking multiple LSTM cells on top of the original structure.
As shown in Figure \ref{fig:deep}, {\it skip connections} are applied to the inputs of all hidden layers as well as between all hidden layers and the outputs \cite{Graves13}.
This reduces the number of processing steps between the bottom of the network and the top, and therefore mitigates the vanishing gradient problem \cite{279181} in the vertical direction. 
To allow all hidden layer information to influence the reading gate, Equation \ref{eq:rgate} is changed to
\begin{equation}
\mathrm{\mathbf{r}}_t = \sigma( \mathrm{\mathbf{W}}_{wr}\mathrm{\mathbf{w}}_t +\sum_{l}\alpha_{l}\mathrm{\mathbf{W}}_{hr}^l\mathrm{\mathbf{h}}_{t-1}^l )
\label{eq:rgatenew}
\end{equation}
where $l$ is the hidden layer index and $\alpha_{l}$ is a layer-wise constant. 
Since the network tends to overfit when the structure becomes more complex, the dropout technique \cite{JMLRsrivastava14a} is used to regularise the network.
As suggested in \cite{ZarembaSV14}, dropout was only applied to the non-recurrent connections, as shown in the Figure \ref{fig:deep}.
It was not applied to word embeddings since pre-trained word vectors were used.

\subsection{Backward LSTM reranking}
\label{sec:brnn}

One remaining problem in the structure described so far is that the LSTM generator selects words based only on the preceding history, whereas some sentence forms depend on the backward context.
Previously, bidirectional networks \cite{schuster1997bidirectional} have been shown to be effective for sequential problems \cite{graves2013speech,D141003}.
However, applying a bidirectional network directly in the SC-LSTM generator is not straightforward since the generation process is sequential in time. 
Hence instead of integrating the bidirectional information into one network, we trained another SC-LSTM from backward context to choose best candidates from the forward generator outputs.
In our experiments, we also found that by tying the keyword detector weights $\mathrm{\mathbf{W}}_{wr}$ (see Equations \ref{eq:rgate} and \ref{eq:rgatenew}) of both the forward and backward networks together makes the generator less sensitive to random initialisation.

\subsection{Training}
\label{sec:train}

The forward generator and the backward reranker were both trained by treating each sentence as a mini-batch.
The objective function was the cross entropy error between the predicted word distribution $\mathrm{\mathbf{p}}_t$ and the actual word distribution $\mathrm{\mathbf{y}}_t$ in the training corpus.
An $l_2$ regularisation term was added to the objective function for every 10 training examples as suggested in \newcite{rnnlm_tool}.
However, further regularisation was required for the reading gate dynamics.  This resulted in the following
modified cost function for each mini-match (ignoring standard $l_2$), 
\begin{equation}
\Scale[0.85]{F(\theta) = \sum_t\mathrm{\mathbf{p}}_t^{\intercal}log(\mathrm{\mathbf{y}}_t)+ \|\mathrm{\mathbf{d}}_T\|+\sum_{t=0}^{T-1}\eta\xi^{\|\mathrm{\mathbf{d}}_{t+1}-\mathrm{\mathbf{d}}_t\|}}
\end{equation}
where $\mathrm{\mathbf{d}}_T$ is the DA vector at the last word index $T$, and $\eta$ and $\xi$ are constants set to $10^{-4}$ and $100$, respectively. 
The second term is used to penalise generated utterances that failed to render all the required slots, while the third term discourages the network from turning more than one gate off in a single time step.
The forward and backward networks were structured to share the same set of word embeddings,  initialised with pre-trained word vectors \cite{pennington2014}.
The hidden layer size was set to be 80 for all cases, and  deep networks were trained with two hidden layers and a 50\% dropout rate.
All costs and gradients were computed and stochastic gradient descent was used to optimise the parameters.
Both networks were trained with back propagation through time \cite{werbos1990backpropagation}.
In order to prevent overfitting, early stopping was implemented using a held-out validation set.

\subsection{Decoding}
\label{sec:decode}

The decoding procedure is split into two phases: (a) over-generation, and (b) reranking.
In the over-generation phase, the forward generator conditioned on the given DA, is used to sequentially generate utterances by random sampling of the predicted next word distributions.
In the reranking phase, the cost of the backward reranker $F_{b}(\theta)$ is computed. 
Together with the cost $F_{f}(\theta)$ from the forward generator, the reranking score $R$ is computed as:
\begin{equation}
\label{eq:score}
R = - ( F_{f}(\theta) + F_{b}(\theta) + \lambda \mathrm{ERR} )
\end{equation}
where $\lambda$ is a tradeoff constant, and the slot error rate $\mathrm{ERR}$ is computed by exact matching the slot tokens in the candidate utterances,
\begin{equation}
\label{eq:sloterr}
 \mathrm{ERR} =  \frac{ p + q }{N}
\end{equation}
where N is the total number of slots in the DA, and $p$, $q$ is the number of missing and redundant slots in the given realisation.
Note that the $\mathrm{ERR}$ reranking criteria cannot handle arbitrary slot-value pairs such as binary slots or slots that take the don't care value because they cannot be delexicalised and exactly matched.
$\lambda$ is set to a large value in order to severely penalise nonsensical outputs.

\section{Experiments}
\label{sec:exp}

\subsection{Experimental Setup}
\label{sec:data}

The target application for our generation system is a spoken dialogue system providing information about certain venues in San Francisco. 
In order to demonstrate the scalability of the proposed method and its performance in different domains, we tested on two domains that talk about restaurants and hotels respectively.
There are 8 system dialogue act types such as {\it inform} to present information about restaurants, {\it confirm} to check that
a slot value has been recognised correctly, and {\it reject} to advise that the user's constraints cannot be met.
Each domain contains 12 attributes (slots), some are common to both domains and the others are domain specific.
The detailed ontologies for the two domains are provided in Table \ref{tab:otg}.
To form a training corpus for each domain, dialogues collected from a previous user trial \cite{gavsic2015distributed} of a statistical dialogue manager were randomly sampled and shown to workers recruited via the Amazon Mechanical Turk (AMT) service. 
Workers were shown each dialogue turn by turn and asked to enter an appropriate system response in natural English corresponding to each system DA.
For each domain around 5K system utterances were collected from about 1K randomly sampled dialogues.
Each categorical value was replaced by a token representing its slot, and slots that appeared multiple times in a DA were merged into one.
After processing and grouping each utterance according to its delexicalised DA, we obtained 248 distinct DAs in the restaurant domain and 164 in the hotel domain.
The average number of slots per DA for each domain is 2.25 and 1.95, respectively.

\begin{table}[t]
\caption{Ontologies used in the experiments.}
\centering 
\vspace{-2 mm}
\setlength{\intextsep}{3pt plus 2pt minus 2pt} 
\setlength{\abovecaptionskip}{0pt}
\setlength{\belowcaptionskip}{0pt}
\label{tab:otg}
\hspace*{0pt}\makebox[\linewidth][c]{%
\scalebox{1.0}{
\begin{tabular}{C{1cm}|C{2.4cm}||C{2.4cm}}
\hline
		&	SF Restaurant 	&	SF Hotel\\
\hline
\parbox[t]{2mm}{{\rotatebox[origin=c]{90}{act type}}}	&\multicolumn{2}{L{4.8cm}}{inform, inform\_only, reject, confirm, select, request, reqmore, goodbye}\\
\hline
\parbox[t]{2mm}{{\rotatebox[origin=c]{90}{shared}}}	&\multicolumn{2}{L{4.8cm}}{name, type, *pricerange, price, phone, address, postcode, *area, *near}\\
\hline
\parbox[t]{2mm}{\multirow{3}{*}{\rotatebox[origin=c]{90}{specific}}}	&*food			&	{\bf *hasinternet}\\
		&*goodformeal		&	{\bf *acceptscards}\\
		&{\bf *kids-allowed}	&	{\bf *dogs-allowed}\\
\hline
\multicolumn{3}{L{7cm}}{ {\small {\bf bold}=binary slots, *=slots can take ``don't care" value}}
\end{tabular}}}
\end{table}

The system was implemented using the Theano library \cite{bergstra2010,Theano2012}, and trained by partitioning each of the collected corpus into a training, validation, and testing set in the ratio 3:1:1. 
The frequency of each action type and slot-value pair differs quite markedly across the corpus, hence up-sampling was used to make the corpus more uniform.
Since our generator works stochastically and the trained networks can differ depending on the initialisation, all the results shown below\footnote{Except human evaluation, in which only one set of networks was used.} were averaged over 5 randomly initialised networks.
For each DA, we over-generated 20 utterances and selected the top 5 realisations after reranking. 
The BLEU-4 metric was used for the objective evaluation \cite{papineni2002bleu}.
Multiple references for each test DA were obtained by mapping them back to the distinct set of DAs, grouping those delexicalised surface forms that have the same DA specification, and then lexicalising those surface forms back to utterances.
In addition, the slot error rate ($\mathrm{ERR}$) as described in Section \ref{sec:decode} was computed as an auxiliary metric alongside the BLEU score.
However, for the experiments it is  computed at the corpus level, by averaging slot errors over each of the top 5 realisations in the entire corpus.
The trade-off weights $\alpha$ between keyword and key phrase detectors as mentioned in Section \ref{subsec:sclstm} and \ref{subsec:deep} were set to 0.5.

\subsection{Objective Evaluation}
\label{sec:objective}

\begin{table}[t]
\caption{Objective evaluation of the top 5 realisations. Except for handcrafted ({\it hdc}) and k-nearest neighbour ({\it kNN}) baselines, all the other approaches ranked their realisations from 20 over-generated candidates. }
\centering 
\vspace{-2 mm}
\setlength{\intextsep}{3pt plus 2pt minus 2pt} 
\setlength{\abovecaptionskip}{0pt}
\setlength{\belowcaptionskip}{0pt}
\label{tab:compare}
\hspace*{0pt}\makebox[\linewidth][c]{%
\scalebox{0.95}{
\begin{tabular}{l|cc||cc}
\Xhline{2\arrayrulewidth}
\multirow{2}{*}{Method}	&	\multicolumn{2}{c||}{SF Restaurant}	&	\multicolumn{2}{c}{SF Hotel}\\
\cline{2-5}
					&	BLEU		&	ERR(\%)		&	BLEU		&	ERR(\%)\\
\Xhline{2\arrayrulewidth}
hdc					&	0.451		&	0.0			&	0.560		& 	0.0 \\
kNN					&	0.602		&	0.87			&	0.676		&	1.87\\
classlm				&	0.627		&	8.70			&	0.734		&	5.35\\
\Xhline{2\arrayrulewidth}
rnn  w/o				&	0.706		&	4.15			&	0.813		&	3.14\\
lstm w/o				&	0.714		&	1.79			&	0.817		&	1.93\\
\Xhline{2\arrayrulewidth}
rnn  w/				&	0.710		&	1.52			&	0.815		&	1.74\\
lstm w/				&	0.717		&	0.63			&	0.818		&	1.53\\
\Xhline{2\arrayrulewidth}
sc-lstm				&	0.711		&	0.62			&	0.802		&	0.78\\
+deep				&	{\bf0.731}		&	{\bf 0.46}		&	{\bf0.832}		&	{\bf 0.41}\\
\Xhline{2\arrayrulewidth}
\end{tabular}
}}
\end{table}

We compared the single layer semantically controlled LSTM ({\it sc-lstm}) and a deep version with two hidden layers  ({\it+deep}) against several baselines: the handcrafted generator ({\it hdc}), k-nearest neighbour ({\it kNN}), class-based LMs ({\it classlm}) as proposed in \newcite{Oh2000}, the heuristic gated RNN as described in \newcite{wenrgm15} and a similar LSTM variant ({\it rnn w/} \& {\it lstm w/}), and the same RNN/LSTM but without  gates ({\it rnn w/o} \& {\it lstm w/o}).
The handcrafted generator  was developed over a long period of time and is the standard generator used for trialling end-to-end dialogue systems (for example \cite{gavsic2014incremental}).
The kNN was implemented by computing the similarity of the test DA 1-hot vector against all of the training DA 1-hot vectors, selecting the nearest
and then lexicalising to generate the final surface form.
The objective results are shown in  Table \ref{tab:compare}.  As can be seen, none of the baseline systems shown in the first block ({\it hdc}, {\it kNN}, \& {\it classlm}) are comparable to the systems described in this paper ({\it sc-lstm} \& {\it+deep}) if both metrics are considered.
Setting aside the difficulty of scaling to large domains, the handcrafted generator's ({\it hdc}) use of predefined rules yields a fixed set of sentence plans, which can differ markedly from the real colloquial human responses collected from AMT, while the class LM approach suffers from inaccurate rendering of information. 
Although the {\it kNN} method provides reasonable adequacy i.e. low $\mathrm{ERR}$, the BLEU is low, probably because of the errors in the collected corpus which {\it kNN} cannot handle but statistical approaches such as LMs can by suppressing unlikely outputs.

\begin{figure*}[t]
\subfloat[An example realisation from SF restaurant domain]{\centerline{\includegraphics[width=165mm]{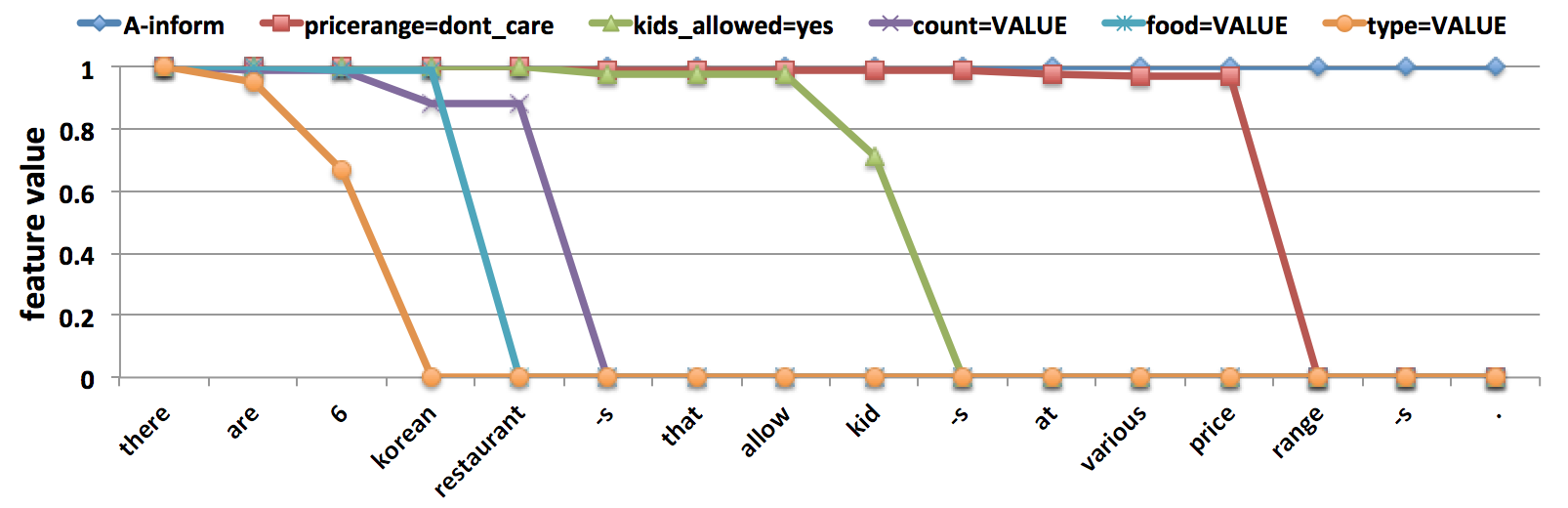}}} \hfill 
\subfloat[An example realisation from SF hotel domain]{	\centerline{\includegraphics[width=165mm]{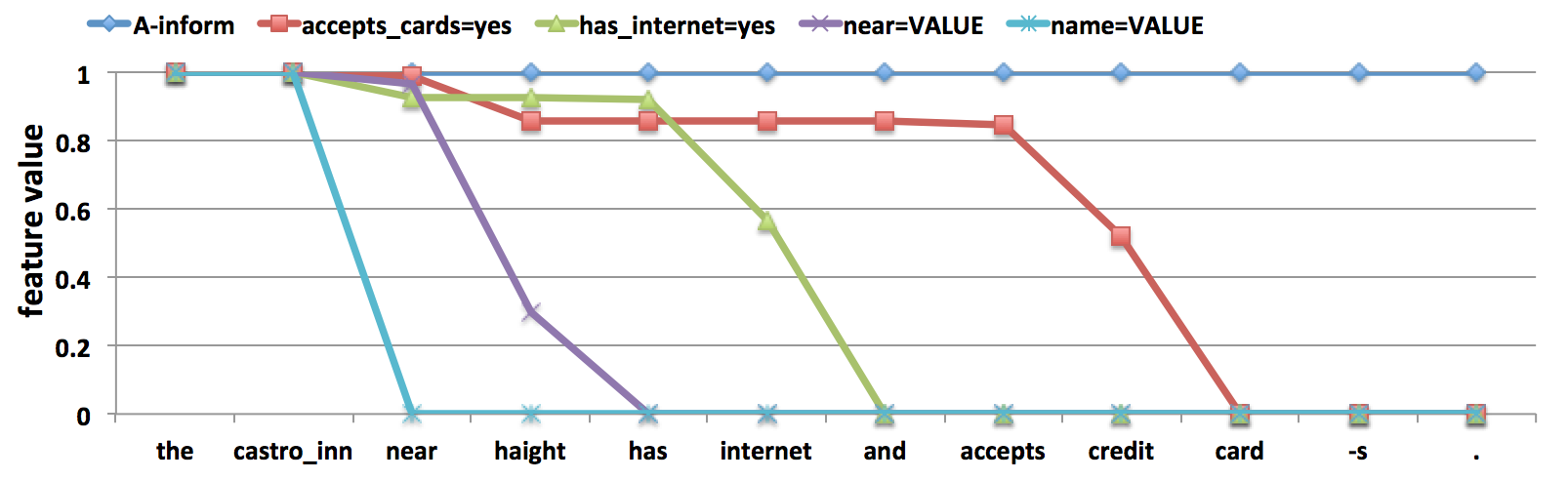}}}
\caption{Examples showing how the SC-LSTM controls the DA features flowing into the network via its learned semantic gates. Despite errors due to sparse training data for some slots, each gate generally learned to detect words and phrases describing a particular slot-value pair. }
\label{fig:example}
\vspace{-2mm}
\end{figure*}

The last three blocks in Table \ref{tab:compare}  compares the proposed method with previous RNN approaches. 
LSTM generally works better than vanilla RNN due to its ability to model long range dependencies more efficiently. 
We also found that by using gates, whether learned or heuristic, gave much lower slot error rates.
As an aside, the ability of the SC-LSTM to learn gates is also exemplified in Figure \ref{fig:example}.
Finally, by combining the learned gate approach with the deep architecture ({\it +deep}), we obtained the best overall performance.

\subsection{Human Evaluation}
\label{sec:human_eval}

\ctable[
  label = tab:qualitytest,
  pos   = t,
  notespar,
  caption={Real user trial for utterance quality assessment on two metrics (rating out of 3), averaging over top 5 realisations. Statistical significance was computed using a two-tailed Student's t-test, between deep and all others. }
]{l c c}
{\tnote[*]{$p<0.05$}\tnote[**]{$p<0.005$}}{
\toprule
\textbf{Method}	& \textbf{Informativeness}	& \textbf{Naturalness}\\ 
\midrule
+deep    		&	2.58		   	 	&      {\bf 2.51}  	 	\\
\midrule
sc-lstm	       	&	{\bf 2.59}			&	2.50			\\
rnn w/    		&	2.53   	 	     	&      2.42\tmark[{\makebox[0pt][l]{*}}]\\
classlm	       	&	2.46\tmark[{\makebox[0pt][l]{**}}]	&	2.45			\\
\bottomrule}

\ctable[
  label = tab:preftest,
  pos   = t,
  notespar,
  caption={Pairwise preference test among four systems. Statistical significance was computed using two-tailed binomial test.}
]{l!{\vrule width 0.8pt}c c c c}
{\tnote[*]{$p<0.05$}\tnote[**]{$p<0.005$}}{
\toprule
{\bf Pref.\%}	&	{\bf classlm}	&	{\bf rnn w/}	&	{\bf sc-lstm}	&	{\bf+deep}\\
\midrule
{\bf classlm}	&	-		&	46.0		&	40.9\tmark[{\makebox[0pt][l]{**}}]	& 	37.7\tmark[{\makebox[0pt][l]{**}}]\\
{\bf rnn w/}	&	54.0		&	-		&	43.0		&	35.7\tmark[{\makebox[0pt][l]{*}}]\\
{\bf sc-lstm}	&	59.1\tmark[{\makebox[0pt][l]{*}}] 	&	57		&	-		&	47.6\\
{\bf+deep}		&	62.3\tmark[{\makebox[0pt][l]{**}}]	&	64.3\tmark[{\makebox[0pt][l]{**}}]	&	52.4		&	-\\
\bottomrule}

Since automatic metrics may not consistently agree with human perception \cite{Stent05evaluatingevaluation}, human testing is needed to assess subjective quality.
To do this, a set of judges were recruited using AMT. 
For each task, two systems among the four ({\it classlm}, {\it rnn w/}, {\it sc-lstm}, and {\it+deep}) were randomly selected to generate utterances from a set of newly sampled dialogues in the restaurant domain. 
In order to evaluate system performance in the presence of language variation, each system generated 5 different surface realisations for each input DA and the human judges were asked to score each of them in terms of informativeness and naturalness (rating out of 3), and also asked to state a preference between the two.
Here {\it informativeness} is defined as whether the utterance contains all the information specified in the DA, and {\it naturalness} is defined as whether the utterance could plausibly have been produced by a human.
In order to decrease the amount of information presented to the judges, utterances that appeared identically in both systems were filtered out.
We tested 1000 DAs in total, and after filtering there were approximately 1300 generated utterances per system.

Table \ref{tab:qualitytest} shows the quality assessments which exhibit the same general trend as the objective results.
The SC-LSTM systems ({\it sc-lstm} \& {\it +deep}) outperform the class-based LMs ({\it classlm}) and the RNN with heuristic gates ({\it rnn w/}) in both metrics.
The deep SC-LSTM system ({\it +deep}) is significantly better than the  class LMs ({\it classlm}) in terms of {\it informativeness}, and better than the RNN with heuristic gates ({\it rnn w/}) in terms of {\it naturalness}.
The preference test results are shown in Table \ref{tab:preftest}.
Again, the SC-LSTM systems ({\it sc-lstm \& +deep}) were significantly preferred by the judges. 
Moreover, the judges recorded a strong preference for the deep approach ({\it +deep}) compared to the others, though the preference is not significant when comparing to its shallow counterpart ({\it sc-lstm}).
Example dialogue acts and their top-5 realisations are shown in Table~\ref{tab:example}. 

\section{Conclusion and Future Work}
\label{sec:conclusion}

\begin{table*}
\centering
\small
\caption{Samples of top 5 realisations from the deep SC-LSTM (+deep) system output. }
\setlength{\intextsep}{3pt plus 2pt minus 2pt} 
\setlength{\abovecaptionskip}{0pt}
\setlength{\belowcaptionskip}{0pt}
\label{tab:example}
\begin{tabular}{|l|l|}
\hline
{\bf\#}& 	{\bf Example Dialogue Acts and Realizations from SF Restaurant Domain}\\
\hline
1	&	inform(name="red door cafe", goodformeal="breakfast", area="cathedral hill", kidsallowed="no")
\\
	&	{\it red door cafe is a good restaurant for breakfast in the area of cathedral hill and does not allow children .}\\
	& 	{\it red door cafe is a good restaurant for breakfast in the cathedral hill area and does not allow children .}\\
	& 	{\it red door cafe is a good restaurant for breakfast in the cathedral hill area and does not allow kids .}\\
	& 	{\it red door cafe is good for breakfast and is in the area of cathedral hill and does not allow children .}\\
	& 	{\it red door cafe does not allow kids and is in the cathedral hill area and is good for breakfast .}\\
\hline
2	& 	informonly(name="dosa on fillmore and kiss seafood", pricerange="expensive", near="lower pacific heights")\\
	&	{\it there is no place other than dosa on fillmore and kiss seafood that are expensive near to lower pacific heights .}\\
	&	{\it dosa on fillmore and kiss seafood is the only expensive restaurant near lower pacific heights .}\\
	&	{\it the only listed restaurant near lower pacific heights in the expensive price range is dosa on fillmore and kiss seafood .}\\
	&	{\it i apologize , dosa on fillmore and kiss seafood is the only expensive restaurant near lower pacific heights .}\\
	&	{\it i apologize , dosa on fillmore and kiss seafood are the only expensive restaurants near lower pacific heights .}\\
\hline
{\bf\#}& 	{\bf Example Dialogue Acts and Realizations from SF Hotel Domain}\\
\hline
3	& 	inform(type="hotel",count="182",dogsallowed="dontcare")\\
	&	{\it there are 182 hotels if you do not care whether dogs are allowed .}\\
	& 	{\it there are 182 hotels if you do not care whether they allow dogs .}\\
	& 	{\it 182 hotels are available if dogs allowed or not is not an issue .}\\
	& 	{\it there are 182 hotels if allowing dogs or not is not an issue .}\\
	& 	{\it there are 182 hotels if whether dogs are allowed does not matter .}\\
\hline
4	& 	informonly(name="red victorian bed breakfast",acceptscreditcards="yes",near="haight",hasinternet="yes")\\
	&	{\it red victorian bed breakfast is the only hotel near haight and accepts credit cards and has internet .}\\
	& 	{\it red victorian bed breakfast is the only hotel near haight and has internet and accepts credit cards .}\\
	& 	{\it red victorian bed breakfast is the only hotel near haight that accept credit cards and offers internet .}\\
	& 	{\it the red victorian bed breakfast has internet and near haight , it does accept credit cards .}\\
	& 	{\it the red victorian bed breakfast is the only hotel near haight that accepts credit cards , and offers internet .}\\
\hline
\end{tabular}
\end{table*}

In this paper we have proposed a neural network-based generator that is capable of generating natural linguistically varied responses based on a deep, semantically controlled LSTM architecture which we call SC-LSTM.
The generator can be trained on unaligned data by jointly optimising its sentence planning and surface realisation components using a simple cross entropy criterion without any heuristics or handcrafting.
We found that the SC-LSTM model achieved the best overall performance on two objective metrics across two different domains.
An evaluation by human judges also confirmed that the SC-LSTM approach is strongly preferred to a variety of existing methods.

This work represents a line of research that tries to model the NLG problem in a unified architecture, whereby the entire model is end-to-end trainable from data.
We contend that this approach can produce more natural responses which are more similar to colloquial styles found in human conversations.
Another key potential advantage of neural network based language processing is the implicit use of distributed representations for words and a single compact parameter encoding of the information to be conveyed.
This suggests that it should be possible to further condition the generator on some dialogue features such discourse information or social cues during the conversation.
Furthermore, adopting a corpus based regime enables domain scalability and multilingual NLG to be achieved with less cost and a shorter lifecycle.
These latter aspects will be the focus of our future work in this area.

\section{Acknowledgements}
Tsung-Hsien Wen and David Vandyke are supported by Toshiba Research Europe Ltd, Cambridge Research Laboratory.

%

\bibliographystyle{acl}
\bibliography{refs}

\end{document}